\documentclass[lettersize,journal]{IEEEtran}
\usepackage{amsmath,amsfonts}
\usepackage{array}
\usepackage[caption=false,font=normalsize,labelfont=sf,textfont=sf]{subfig}
\usepackage{textcomp}
\usepackage{stfloats}
\usepackage{url}
\usepackage{verbatim}
\usepackage{graphicx}
\usepackage{cite}

\usepackage{url}
\usepackage{amsmath,amssymb,amsfonts}
\usepackage{array}
\usepackage{multirow}
\usepackage{amssymb}
\usepackage{booktabs}
\usepackage{hyperref}       
\usepackage{comment}
\usepackage{color}
\usepackage{colortbl}
\usepackage{tikz}
\usepackage{bm}
\usepackage[ruled,linesnumbered]{algorithm2e}

\begin{document}

\title{Detecting Deepfake by Creating Spatio-Temporal Regularity Disruption}
\author{Jiazhi~Guan, Hang~Zhou, Mingming~Gong, Errui~Ding, Jingdong~Wang, Youjian~Zhao}
\maketitle
\begin{abstract}
Despite encouraging progress in deepfake detection, generalization to unseen forgery types remains a significant challenge due to the limited forgery clues explored during training. 
In contrast, we notice a common phenomenon in deepfake: fake video creation inevitably disrupts the statistical regularity in original videos. Inspired by this observation, we propose to boost the generalization of deepfake detection by distinguishing the ``\textit{regularity disruption}'' that does not appear in real videos.
Specifically, by carefully examining the spatial and temporal properties, we propose to disrupt a real video through a \emph{Pseudo-fake Generator} and create a wide range of \emph{pseudo-fake} videos for training. 
Such practice allows us to achieve deepfake detection without using fake videos and improves the generalization ability in a simple and efficient manner.
To jointly capture the spatial and temporal disruptions, we propose a Spatio-Temporal Enhancement block to learn the regularity disruption across space and time on our self-created videos.  
Through comprehensive experiments, our method exhibits excellent performance on several datasets.
\end{abstract}

\begin{IEEEkeywords}
Deepfake Detection, Digital Forensics
\end{IEEEkeywords}
\section{Introduction}
In recent years, our social platforms have been flooded with high-quality manipulated or forged portrait videos. These videos are created by advanced generators \cite{8718508,kim2018deep,suwajanakorn2017synthesizing,xu2021multi,cao2021unifacegan,shao2021explicit,9462525}, known as \textbf{deepfakes}, and often target at celebrities or political figures with misleading or seditious intentions, which can lead to potential misperception of the facts and even social unrest. 
Thus, the detection of deepfakes has become an important topic in the computer vision and related research communities.

With recent efforts devoted to identifying the differences between the real and fake videos~\cite{bayar2016deep,sun2021dual,rahmouni2017distinguishing,luo2021generalizing,chai2020makes,zhao2021multi,das2021towards,haliassos2021lips,gu2021spatiotemporal,sun2021improving}, plausible performance has been achieved on popular datasets \cite{rossler2019faceforensics++,jiang2020deeperforensics}. However, the generalizability of deepfake detection methods remains one of the biggest concerns in the community. 
To address this challenging problem, \cite{wang2021representative} propose a dynamic masking strategy to avoid overfitting on obvious artifacts. With similar intuitions, \cite{wang2022deepfake,chen2022self} propose to boost the generalization by adversarial training. Moreover, \cite{zheng2021exploring,masi2020two} consider the generalizable temporal coherence, and \cite{luo2021generalizing,li2021frequency,qian2020thinking} try to discover more subtle clues in the frequency domain. 
While their designs achieved better results, these methods leverage only existing datasets generated by limited forgery techniques. As a result, they inevitably tend to recognize only a subset of deepfake's ``fingerprint". 
Different from the dataset-dependent methods, \cite{li2018exposing,li2020face} propose to imitate the deepfake pipeline and expect their models to learn the discriminatory factors depending on face swapping or warping traces.
However, their models still focus on limited artifacts, thus leading to sub-optimal generalization.

\begin{figure*}[!t]
\centering
\includegraphics[width=\linewidth]{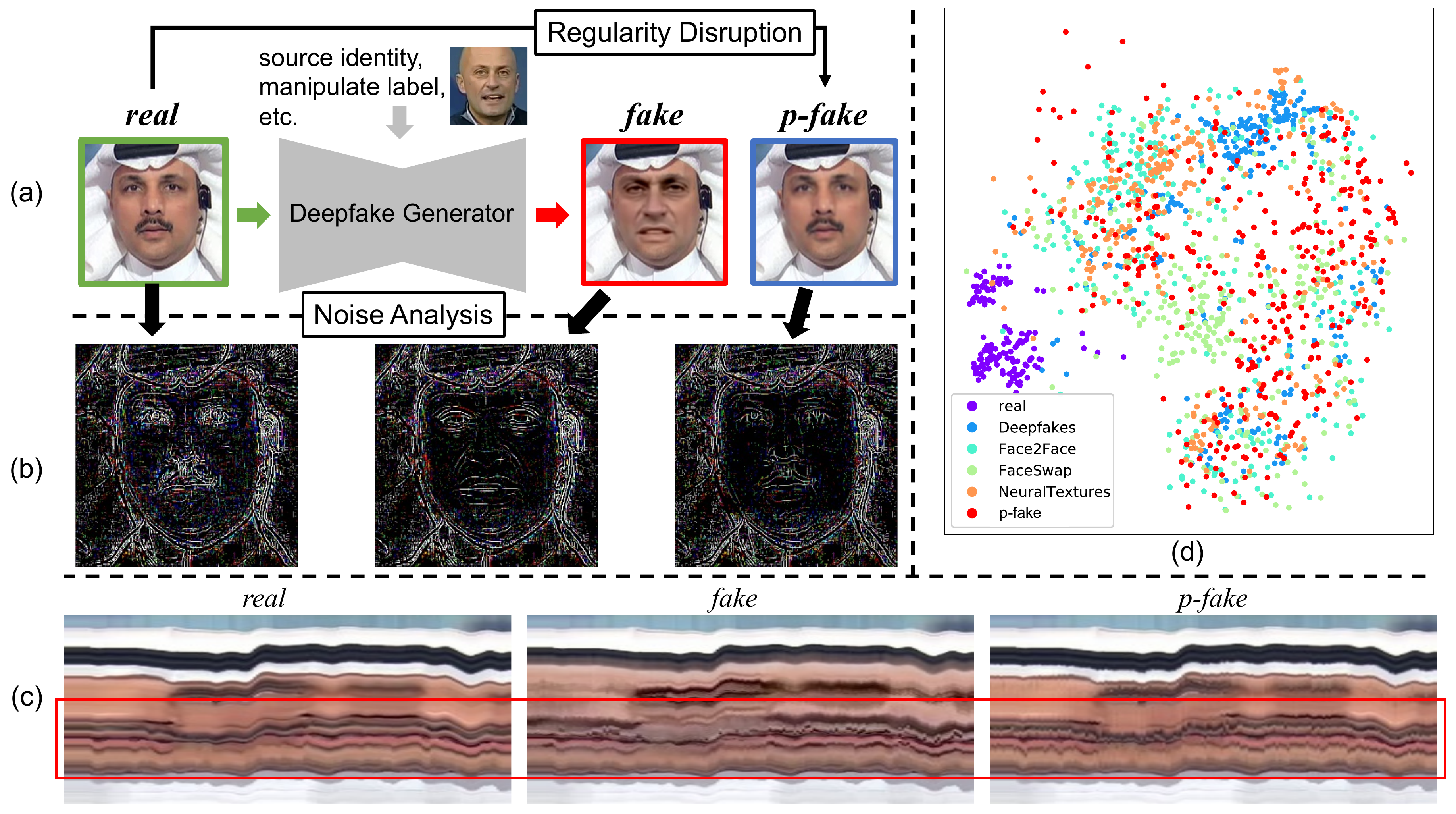}
\caption{
\textbf{(a)}: Differences between the three kinds of data, \textit{p-fake} keeps the identity and facial movements of the original \textit{real} one, while \textit{fake} is generated by forgery techniques with modified identity or facial movements. \textbf{(b)}: Noise analysis of three kinds of data, where our \textit{p-fake} simulates the noise regularity disruption. \textcolor{black}{It can be observed that the facial region displays smoother textures.} \textbf{(c)}: Temporal visualization of three kinds of data, where the \textit{real} one shows smoother fluctuation, and our created \textit{p-fake} demonstrates the similar temporal irregularity as the \textit{fake}. \textbf{(d)}: The T-SNE visualization of features extracted by our model trained using only \textit{real} and \textit{p-fake} data, each dot corresponds to the feature of one test video}
\label{fig:fig1}
\end{figure*}

One shared intuition among most of previous methods is that they pursue to cover the forgery clues caused by producing deepfakes, such as generative artifacts and evident post-processing traces.
In contrast to previous methods, we do not focus on visible and explicit deepfake fingerprints, but rather on the underlying statistical properties of \emph{{real}} and \emph{{fake}} videos, which we refer to as ``regularities''. A few important observations are firstly made: \textbf{1)} The creation of \emph{\textbf{fake}} videos inevitably requires modifying and re-assembling the facial parts of the \emph{\textbf{real}} videos, regardless of the deepfake technique.  \textbf{2)} Spatial modifications would break the homogeneity of images' statistical properties \cite{farid2009image}, which are usually introduced by the imaging~\cite{chen2011lens,gonzalez2007digital} and image/video compression~\cite{wallace1992jpeg,richardson2011h} pipelines. We show a specific noise analysis example in Fig.~\ref{fig:fig1}~(a), and (b).  \textbf{3)} The original temporal continuity of the \textit{real} videos is destroyed when re-assembling separately generated deepfake frames. As shown in Fig.~\ref{fig:fig1}~(c), we horizontally stitch the vertical slices of all frames in a video in the same column. The \textit{real} video shows smoother fluctuation than the \textit{fake} one. We uniformly name the observed destruction of homogeneity in videos ``\emph{regularity disruption}''.

Guided by the observations above, we anchor our solution on creating {\textit{pseduo-fake}} (\emph{\textbf{p-fake}}) data to imitate both spatial and temporal \emph{regularity disruption}.  The model trained to distinguish \textit{real} from our self-generated \textit{p-fake} data should naturally be suitable for the task of deepfake detection once enough \emph{regularity disruption} distributions can be covered.
To this end, we propose a plug-and-play module named {Pseudo-fake Generator} (\textbf{\textit{P-fake} Generator}).
Spatially, we introduce the regularity disruption through our elaborately designed blend-after-edit pipeline, in which various kinds of edits are considered at photometric perspective, geometric perspective, and frequency domain.
Temporally, frames are processed individually with random parameters to break the temporal regularity.
Fig.~\ref{fig:fig1} (a)(b)(c) show an example where our \textit{p-fake} reproduces the irregularities in a very similar manner as the \textit{fake} results. In addition, as shown in the T-SNE \cite{van2008visualizing} visualization in Fig.~\ref{fig:fig1}~(d), the generated \textit{p-fake} covers a broad range of irregularity features.

Based on our \textit{P-fake} generator,  we present a specially devised Spatio-Temporal Enhancement (\textbf{STE}) block to jointly consider spatio-temporal \emph{regularity disruption}. 
In STE, channel-wise temporal convolution is first applied to achieve adjacent interactions. Then, by sequentially plugging self-attention spatial re-weighting, the STE block yields spatio-temporally enhanced encodings.

The contributions of this paper are summarized as follows:

\textbf{1)} 
For the first time, we propose to detect deepfake supported by the introduced \textit{p-fake} with self-created spatio-temporal regularity disruptions.

\textbf{2)} To generate desirable \textit{p-fake}, we propose a plug-and-play module, \textit{P-fake} Generator, which adds zero computation to the detector but greatly promotes the performance. Furthermore, the proposed \textit{P-fake} Generator can be applied to all existing learning-based deepfake detection methods.

\textbf{3)} By jointly considering the spatial and temporal regularity disruptions, we propose the Spatio-Temporal Enhancement block to better capture the irregular patterns.

\textbf{4)} Through comprehensive experiments, our method exhibits excellent performance in both in-dataset and cross-dataset settings. Especially, the generalization performances are improved by 10.67 in terms of AUC\% on the challenging Deepwild dataset~\cite{zi2020wilddeepfake}.

\section{Related Works}
Media forensics has been extensively studied \cite{farid2009image,verdoliva2020media,9715146,8626149} for a long time, many efforts are devoted to detect specific types of manipulation, such as splicing \cite{huh2018fighting,liu2019adversarial}, copy-move \cite{liu2021two,7517389,7086315}, and object removal \cite{9089005,stamm2012temporal}.
Although promising results are delivered, these methods are hard to deal with the emerging deepfakes since deepfakes are often generated by advanced GAN-based methods without explicitly copying, pasting, removal, etc.  

Deepfake detection is turning out to be a very hot research topic, as high-quality deepfakes are seriously impairing the dissemination of real information in the Internet age. 
Early works \cite{yang2019exposing,agarwal2019protecting,li2018ictu} are devoted to identifying deepfake by its obvious artifacts. Li~et~al. \cite{li2018ictu} notice the abnormal eye blink signal that is not well presented in early deepfakes. Yang~et~al. \cite{yang2019exposing} consider the inconsistency of head pose across the central and whole face. Despite promising performances at the time, it is difficult today to rely on these apparent forgery artifacts to counter the increasingly sophisticated deepfake techniques.

As the deepfakes become photo-realistic, recent efforts are devoted to exploring more subtle forgery clues. Dang~et~al. \cite{dang2020detection} propose to detect deepfakes with an attention mechanism, which highlights the forgery regions to support classification. Zhao~et~al. \cite{zhao2021multi} develop a multi-attention framework with the similar idea to enrich the shallow feature encoding. Moreover, without a complex model, Das~et~al. \cite{das2021towards} take a simple but efficient way to augment the training data by randomly masking parts of facial features.
Combining with the state-of-the-art (SOTA) CNN backbone, they won the first prize in the most famous deepfake detection challenge~\cite{dolhansky2020deepfake}. 
A related idea is studied in \cite{wang2021representative}, which further takes a dynamic masking processing to suppress overfitting on obvious artifacts.
In addition to the RGB domain, many works \cite{qian2020thinking,masi2020two,luo2021generalizing,liu2021spatial,li2021frequency} notice that forgery clues are more discriminative in the frequency domain. Qian~et~al. \cite{qian2020thinking} explore both the local and global frequency representations of deepfakes. Masi~et~al. \cite{masi2020two} devise frequency encoding into their two-stream model using the Laplacian of Gaussian operator \cite{burt1987laplacian}. Although frequency features show remarkable promotion, it is foreseeable that defects in the frequency domain will be less conspicuous, just like in the RGB domain.

To boost the generalization, another commonly applied method is adversarial training. Wang et al. \cite{wang2022deepfake} propose a blurring-based adversarial training mechanism and a GAN-based generator, where the former brings the detector better robustness to visual compression and latter acts as a surrogate deepfake model in training. Also with adversarial training, a recent work of Chen et al. \cite{chen2022self} additionally restrict the detector to recognize the forgery type as well as the authenticity. These efforts introduce a trainable surrogate deepfake model in the training pipeline to enhance the diversity of deepfakes, which may be able to create deepfakes with fewer visual artifacts, but still cannot jump out of the arms race between detectors and forgeries, since untapped generators always produce different ``fingerprint''.

When video data are available, temporal features are also considered in recent works \cite{haliassos2021lips,qi2020deeprhythm,ganiyusufoglu2020spatio,lu2021deepfake,gu2021spatiotemporal,zheng2021exploring}. 
\textcolor{black}{Combined with a multiple-instance learning framework, Li et al. \cite{li2020sharp} investigate the temporal features using basic 1D-convolutions.}
In \cite{qi2020deeprhythm}, PPG signal \cite{yu2019remote}, which measures the minuscule periodic changes of skin color due to blood pumping, is adopted to identify deepfakes. Haliassos~et~al. 
\cite{haliassos2021lips} propose to utilize the high-level semantic irregularities in mouth movements for a more generalizable deepfake detection. More generally, the temporal inconsistency of deepfakes can be reflected in more low-level features. In a data-driven way, Ganiyusufoglu et al. \cite{ganiyusufoglu2020spatio} adopt the 3D convolutional neural networks to achieve deepfake detection. 
Based on the 3D convolution, Lu~et~al. \cite{lu2021deepfake} propose a 3D-attention mechanism to further improve the performance. 
To avoid spatial overfitting, Zheng~et~al. \cite{zheng2021exploring} propose a fully temporal convolutional network to enhance the generalization capability.

The closest works to us \cite{li2018exposing,li2020face,zhao2021learning} imitate the deepfake pipeline and try to identify deepfakes depending on face swapping or warping traces.
Though we also create negative samples for training using only \textit{real} data, the most essential differences are twofold: 
1) they propose a data generation pipeline to create deepfake artifacts, which are not substantially different from using existing datasets, while we are not intended to create artifacts, but to break the regularities of \textit{real} videos.
2) we anchor our solution to the fundamental regularity disruption in both spatial and temporal spaces, which is more generalizable to various kinds of deepfakes. 
We notice the concurrent work of \cite{shiohara2022detecting} also proposes a similar self-blending pipeline to create more hardly recognizable forgery traces for better generalization. Different from \cite{shiohara2022detecting}, we draw inspiration from a spatio-temporal regularity perspective that does not capitalise on designated kind of forgery trace but on the overall inconsistency (which is intuitively validated in Fig.~\ref{fig:fig5}).

\section{Proposed Method}
In this section, we first introduce our proposed \textit{Pseudo-fake} Generator, then elaborate on the Spatio-Temporal Enhancement block, and finally explain how to train the deepfake detector with the proposed \textit{Pseudo-fake} Generator.
\subsection{Pseudo-fake Generator}
\label{sec:P-fake Generation}

\begin{figure*}[!th]
\centering
\includegraphics[width=\linewidth]{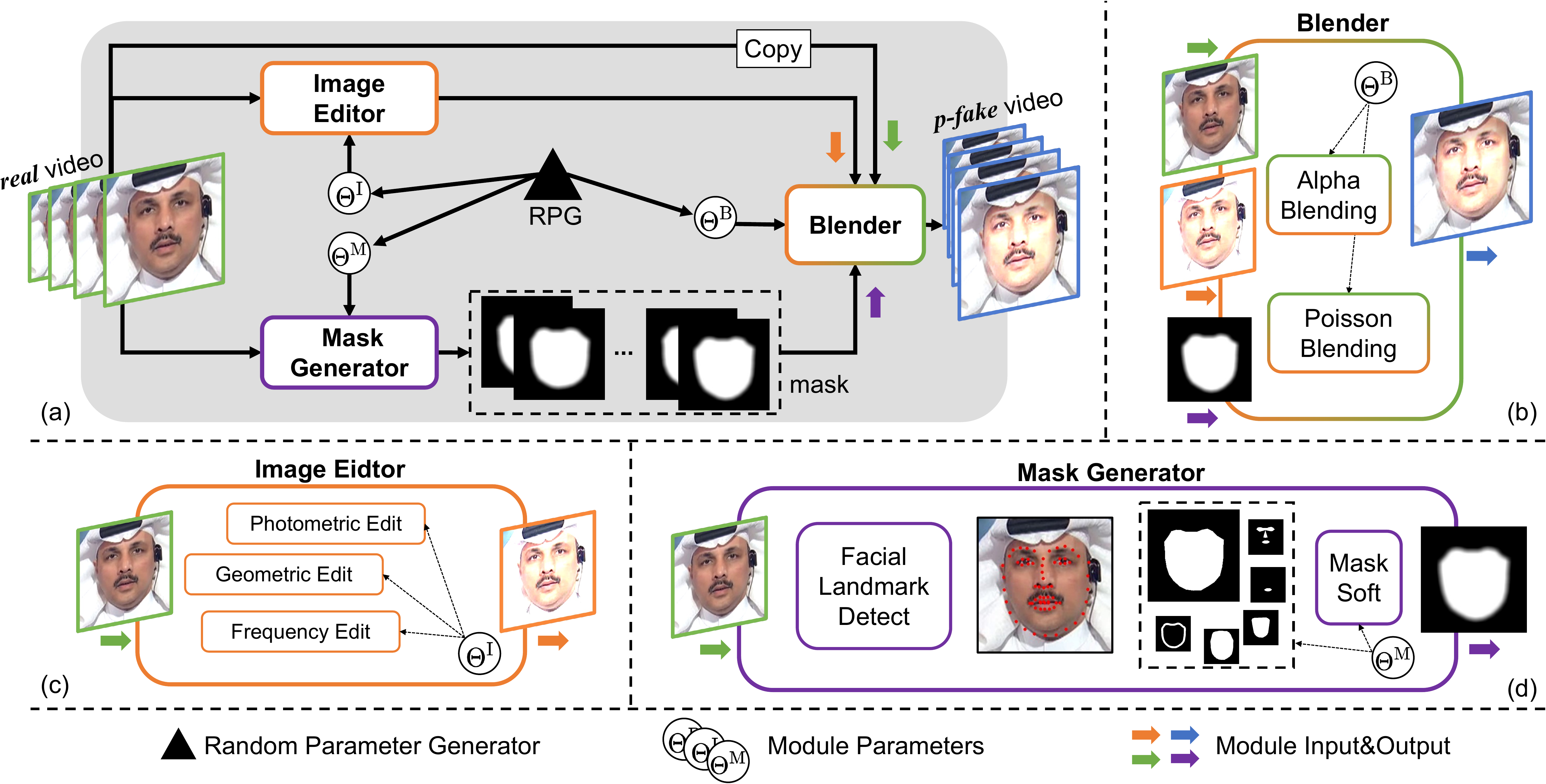}
\caption{Overview of the \textit{P-fake} Generator. We illustrate the main processing in \textbf{(a)}.
The ${\bm{\mathrm{Image\ Editor}}}$ is responsible for editing frames, and the ${\bm{\mathrm{Blender}}}$ will blend parts of these disruptions into the \textit{real} frame with the mask provided by the ${\bm{\mathrm{Mask\ Generator}}}$ to create \textit{p-fake}.
Frames in one video are processed individually with random parameters given by the RPG to break the temporal regularity. 
More details of the three modules are illustrated in \textbf{(b)}, \textbf{(c)}, and \textbf{(d)}.}
\label{fig:fig2}
\end{figure*}

As mentioned in our motivation, we aim to create \textit{p-fake} with the fundamental regularity disruption, thus allowing our model to learn more generalized discriminatory features. To this end, we propose the \textit{P-fake} Generator to create \textit{p-fake} using only the \textit{real} video. We illustrate the main processing of the \textit{P-fake} Generator in Fig.~\ref{fig:fig2}~(a) with a overall description in Algorithm~\ref{alg:pfake_generator} and depict more details about the three modules in Fig.~\ref{fig:fig2}~(b), (c), (d), respectively. Given a \textit{real} video, our \textit{P-fake} Generator handles it frame by frame, where the $\hyperref[sec:method_image_editor]{\bm{\mathrm{Image\ Editor}}}$ is responsible for editing frames, and the $\hyperref[sec:method_mask_g_blender]{\bm{\mathrm{Blender}}}$ will blend parts of these disruptions into the \textit{real} frame with the mask provided by the $\hyperref[sec:method_mask_g_blender]{\bm{\mathrm{Mask\ Generator}}}$ to create \textit{p-fake}. In addition, there is a Random Parameter Generator ($\hyperref[sec:method_rpg]{\bm{\mathrm{RPG}}}$) that plays an important role in the whole process by controlling the diversity of disruption generation, injection regions, and blending methods over the three modules. 
{\color{black} 
For all our notations in follows, we uniformly use subscripts for indices and superscripts to distinguish variables.
}

\begin{algorithm}[!b]
\caption{Pseudo-fake Generator.}
\label{alg:pfake_generator}
\KwIn{
~real video clip $\bm{\mathrm{v}}^{real} \in \mathbb{R}^{L\times H\times W\times C}$, \\
\qquad\quad~ facial landmarks $\bm{\mathrm{f}} \in \mathbb{R}^{L\times 68\times 2}$.
}
\KwOut{pseudo-fake video clip $\bm{\mathrm{v}}^{\operatorname{p-fake}} \in \mathbb{R}^{L\times H\times W\times C}$.}
\For{$t\leftarrow 1$ \KwTo $L$}
{
$I^{real}\in \mathbb{R}^{H\times W\times C} \leftarrow \bm{\mathrm{v}}^{real}_t$\;
$f\in \mathbb{R}^{68\times 2} \leftarrow \bm{\mathrm{f}}_t$\;
\tcp{generate random parameters by \textbf{RPG} for each frame}
$\mathrm{\Theta^I}, \mathrm{\Theta^M}, \mathrm{\Theta^B} \leftarrow \hyperref[sec:method_rpg]{\bm{\mathrm{RPG}}}$\;
\tcp{edit and blend}
{\color{black}
$I^{edited}\leftarrow \hyperref[sec:method_image_editor]{\bm{\mathrm{Image\ Editor}}}(I^{real};~\mathrm{\Theta^I})$\;
$I^{mask}\leftarrow \hyperref[sec:method_mask_g_blender]{\bm{\mathrm{Mask\ Generator}}}({f};~\mathrm{\Theta^M})$\;
$I^{\operatorname{p-fake}}\leftarrow \hyperref[sec:method_mask_g_blender]{\bm{\mathrm{Blender}}}(I^{real}, I^{edited}, I^{mask};~\mathrm{\Theta^B})$\;
}
$\bm{\mathrm{v}}^{\operatorname{p-fake}}_t\leftarrow 
I^{\operatorname{p-fake}}$\;
}
\end{algorithm}

\subsubsection{Image Editor} 
\label{sec:method_image_editor}
Given a \textit{real} frame $I^{real} \in \mathbb{R}^{H\times W\times 3}$, we consider the possible editing methods from three aspects. 

\noindent\textbf{Photometric Perspective.}
We apply several editing methods: \textit{ISO noise}, \textit{sharpen}, \textit{downsampling}, and \textit{color jitter}, to modify the original statistic properties of the \textit{real} frame. Here we present one of them in detail.
For \textit{color jitter}, it is used to alter the original brightness, contrast, and saturation of the \textit{real} frame. We first define a mapping table 
{\color{black}
$\mathrm{T} = [0,1,\dots,255]\in \mathbb{R}^{256}$
}
and three random parameters $\theta^b$, $\theta^t$, $\theta^a$. 
The brightness adjustment is implemented as:
{\color{black}
\begin{equation}
\mathrm{T}^b_k = \mathrm{min}(\mathrm{max}(\left \lfloor \mathrm{T}_k\cdot \theta^b \right \rfloor , 0), 255),~0\le k<256,
\end{equation}
\begin{equation}
I^{edited}_{i,j,c} = \mathrm{T}^b_{I^{real}_{i,j,c}},~0\le i<H,~0\le j<W,~0\le c<3.
\end{equation}
}
In a similar manner, contrast adjustment can be learned by:
{\color{black}
\begin{equation}
\mu=\frac{1}{H\times W}\sum\mathrm{RGB2GRAY}(I^{real}),
\label{eq:RGB2GRAY_ref}
\end{equation}
\begin{equation}
\mathrm{T}^s_k = \mathrm{min}(\mathrm{max}(\left \lfloor \mathrm{T}_k\cdot \theta^t + \mu \cdot (1-\theta^t) \right \rfloor, 0), 255),
\end{equation}
where $0\le k<256$,
\begin{equation}
I^{edited}_{i,j,c} = \mathrm{T}^s_{I^{real}_{i,j,c}},~0\le i<H,~0\le j<W,~0\le c<3.
\end{equation}
Furthermore, saturation adjustment is implemented at each pixel by:
\begin{equation}
\begin{aligned}
I^{edited}_{i,j,c} = &\mathrm{min}(\mathrm{max}(\lfloor I^{real}_{i,j,c}\cdot\theta^a\\
&+\mathrm{RGB2GRAY}(I^{real})_{i,j}\cdot(1-\theta^a) \rfloor, 0), 255).
\end{aligned}
\label{eq:saturation}
\end{equation}
Note $\mathrm{RGB2GRAY}$ in Eq.~\eqref{eq:RGB2GRAY_ref}, \eqref{eq:saturation} denotes the transformation from RGB color space to gray scale, which is implemented by weighted average from RGB channels.
}

\begin{figure*}[!b]
\centering
\includegraphics[width=0.9\linewidth]{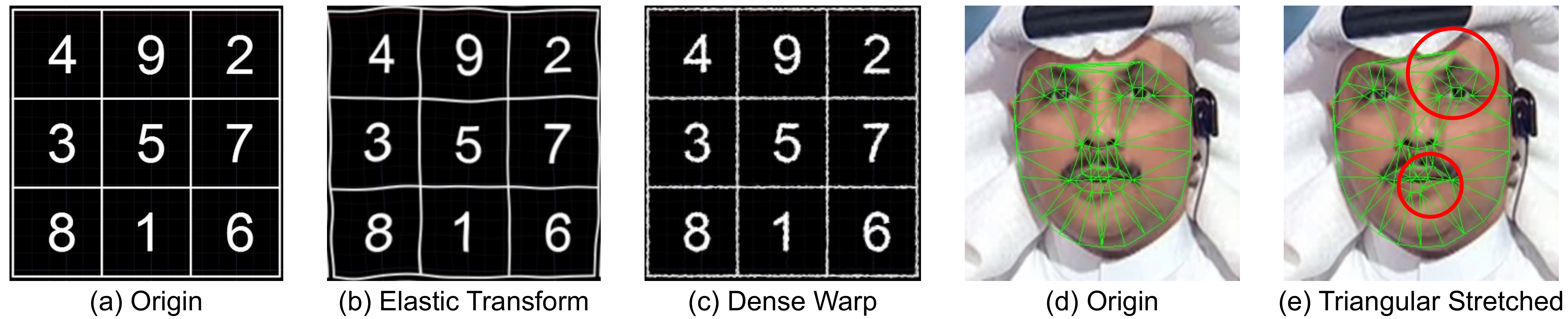}
\caption{Examples of geometric editing results. (a) and (d) are two original images, (b), (c), and (e) present three kinds of editing.}
\label{fig:supp_f1}
\end{figure*}

\noindent\textbf{Geometric Perspective.}
We also introduce different ways to disrupt the original regularity including \textit{elastic transform}, \textit{dense warp} and \textit{triangular stretch}. We elaborate on \textit{elastic transform} here and illustrate some examples for better understanding in Fig~\ref{fig:supp_f1}.

Specifically, the \textit{elastic transform} is implemented by per-pixel displacement to disrupt the original spatial relations. We first create two grid-point matrices as:
{\color{black}
\begin{equation}
\mathrm{G^x}=[\underbrace{~[0,1,\cdots,W-1]^{\mathrm{T}},~\cdots}_{[0,1,\cdots,W-1]^{\mathrm{T}}~\mathrm{repeat}~H~\mathrm{times} }],
\end{equation}
\begin{equation}
\mathrm{G^y}=[\underbrace{~[0,1,\cdots,H-1],~\cdots}_{[0,1,\cdots,H-1]~\mathrm{repeat}~W~\mathrm{times} }]^{\mathrm{T}}.
\end{equation}
}
Then, by random sampling and Gaussian blur, we have the disruption matrix $\mathrm{\Delta'}=\mathrm{GaussianBlur}(\mathrm{\Delta}, \theta^{\sigma})\cdot \theta^{\alpha}$, where $\mathrm{\Delta}\in \mathbb{R}^{H\times W}$ is sampled from the uniform distribution $\mathcal{U}(-1,1)$, and $\theta^{\sigma}$, $\theta^{\alpha}$ are two parameters given by the RPG. After that, the disruption matrix is added to the grid-point matrix, we have 
{\color{black}
$\mathrm{{G^x}'}=\mathrm{max}(\lfloor \mathrm{G^x}+\mathrm{\Delta'} \rfloor, 0)$. 
}
Meanwhile, $\mathrm{{G^y}'}$ is calculated in the same way. As a result, the edited frame is given by:
{\color{black}
\begin{equation}
I^{edited}_{i,j,c} = I^{real}_{\mathrm{G}^\mathrm{x'}_{i,j}, \mathrm{G}^\mathrm{y'}_{i,j}, c},
\label{cv2_remap}
\end{equation}
where $~0\le i<H,~0\le j<W,~0\le c<3$.
}

\noindent\textbf{Frequency Domain.}
{\color{black}
Recent studies \cite{qian2020thinking,luo2021generalizing,liu2021spatial,li2021frequency,gu2022exploiting} have found that inconsistencies hidden in the frequency domain can be used to reliably distinguish deepfakes. Discrete Cosine Transform (DCT) \cite{ahmed1974discrete}, which is widely used in image processing, such as JPEG and H.264, can be more compatible and efficient with the description of compression artifacts that are present in forgery patterns \cite{qian2020thinking,li2021frequency,gu2022exploiting}.

Recent studies \cite{qian2020thinking,luo2021generalizing,liu2021spatial,li2021frequency,gu2022exploiting} have indicated that concealed irregularities in the frequency domain can be utilized to reliably identify deepfakes. Discrete Cosine Transform (DCT) \cite{ahmed1974discrete}, a commonly used technique in image processing, such as JPEG and H.264, can be a more effective and adaptable method to describe the compression artifacts that appear in forgery patterns \cite{qian2020thinking,li2021frequency,gu2022exploiting}.
We thus first disrupt the original DCT responses to simulate inconsistencies in the frequency domain,
}
and then apply Inversed Discrete Cosine Transform (IDCT) to recover the RGB representation. 
This processing is demonstrated as:
{\color{black}
\begin{equation}
I^{edited}=\mathcal{D}^{-1} \left ( \mathcal{D} \left ( I^{real} \right ) + 2*\mathrm{Sigmoid}(\mathrm{\Delta^{F}})-1 \right ), \label{eq:dct_trans_1}
\end{equation}
where $\mathcal{D}$ indicates applying Discrete Cosine Transform \cite{ahmed1974discrete}, $\mathcal{D}^{-1}$ indicates applying Inversed Discrete Cosine Transform, $\mathrm{\Delta^F} \in \mathbb{R}^{H\times W}$ is the noise matrix sampled from the normal distribution as $\mathrm{\Delta^{F}} \sim \mathcal{N}(0,1+\theta^{f})$, and $\theta^{f}$ is a random parameter given by the RPG. Note in Eq.~\eqref{eq:dct_trans_1}, we edit each channel of the RGB input separately, omitting the subscripts for brevity.
}

{\color{black}
\noindent\textbf{Combinations of Editing Methods.}
For all the editing methods, we introduce them with the input denoted $I^{real}$ and output denoted $I^{edited}$. During generation, each of the methods has an equal probability of 30\% of being executed, with the exception of \textit{color jitter}, which has a guaranteed probability of 100\% in order to ensure that at least one type of disruption is present in the output. 
Consequently, different combinations of editing methods result in various forms of disruptions.
}

\subsubsection{Mask Generator and Blender}
\label{sec:method_mask_g_blender}
\begin{figure}[!b]
\centering
\includegraphics[width=0.8\linewidth]{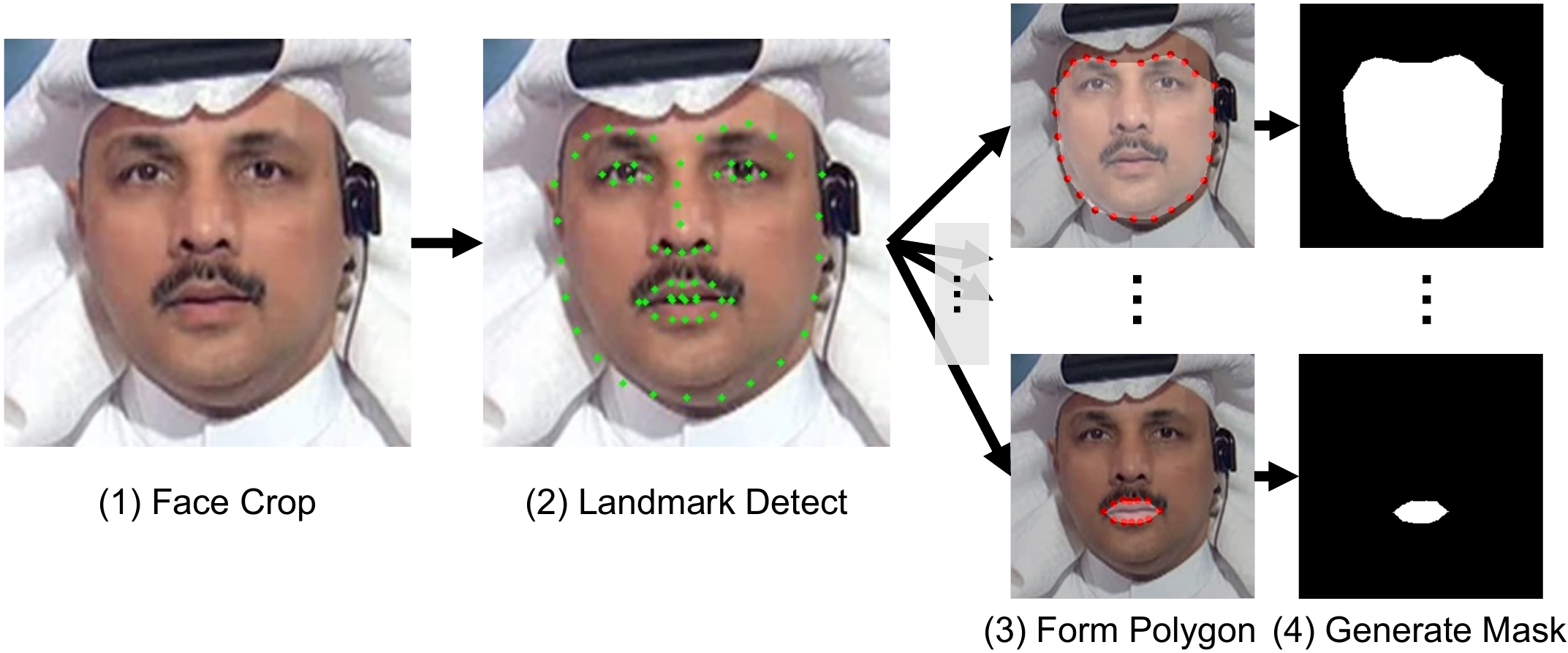}
\caption{
{\color{black}
Mask generation: (1) The face region is cropped from the input frame. (2) We detect 68-point facial landmarks using an off-the-shelf tool. (3)\&(4) We select desired points to form a polygon that covers the region to be masked.
}
}
\label{fig:fig6}
\end{figure}

After the Image Editor creates edited frames, we further blend these disruptions into parts of the \textit{real} frame with Mask Generator and Blender to create irregularities in \textit{p-fake}. 
{\color{black} 
Considering modifications of deepfakes at different regions, we adopt different masks in Fig.~\ref{fig:fig2} (d), including whole face, narrowed face, face with forehead, face boundary, mouth region, facial organs.
The use of different masks results in a generation of \textit{p-fake} with disruptions featured by various spatial patterns, which greatly enriches the diversity of \textit{p-fake}.
In our Mask Generator, we first extract the facial landmarks, then generate kinds of masks by joining the landmarks to form a closed polygon. For example, generation of the ``whole face'' and ``mouth region'' masks is demonstrated in Fig.~\ref{fig:fig6}. Other masks are generated using a similar method, with different polygons formed to cover the desired region.
uring generation, we randomly adopt a mask from \{whole face, narrowed face, face with forehead\} with a probability of 0.75, while from \{face boundary, mouth region, facial organs\} with a probability of 0.25.
This stochastic selection is also controlled by the RPG.
}
After that, the mask is deformed using elastic transform as introduced earlier, but such practice serves a different purpose of mask augmentation. Then, the mask is softened using Gaussian blur with a randomly chosen kernel size $\theta^k$. We also randomly soften the background or foreground to create inconsistent overall smoothness after blending.
After that, we stitch the \textit{real} and edited frames in the certain mask region using randomly chosen blending method (see Fig.~\ref{fig:fig2}~(b)). 
Taking alpha blending as an example, the \textit{p-fake} is finally obtained as: $I^{\operatorname{p-fake}}=I^{real}\cdot (1-I^{mask}) + I^{edited} \cdot I^{mask}$, where $I^{\operatorname{p-fake}}$ fuses $I^{real}$ and $I^{edited}$ together to represent regularity disruption.

\subsubsection{RPG}
\label{sec:method_rpg}
RPG is the key to creating temporal irregularity. As described above, during the \textit{p-fake} generation, RPG provides parameters ($\theta^{\star}$) to affect the final result.
Before processing each frame, RPG will sample a set of parameters from certain ranges uniformly. In this way, frames in a video are given a distinct set of parameters, allowing different frames to be edited to different degrees or by different combinations of methods, resulting in temporal regularity disruption. To prevent frequent temporal changes, we ensure that the parameters between randomly selected consecutive frames remain the same. 

\subsection{Spatio-Temporal Enhancement}
\begin{figure*}[!th]
\centering
\includegraphics[width=0.9\linewidth]{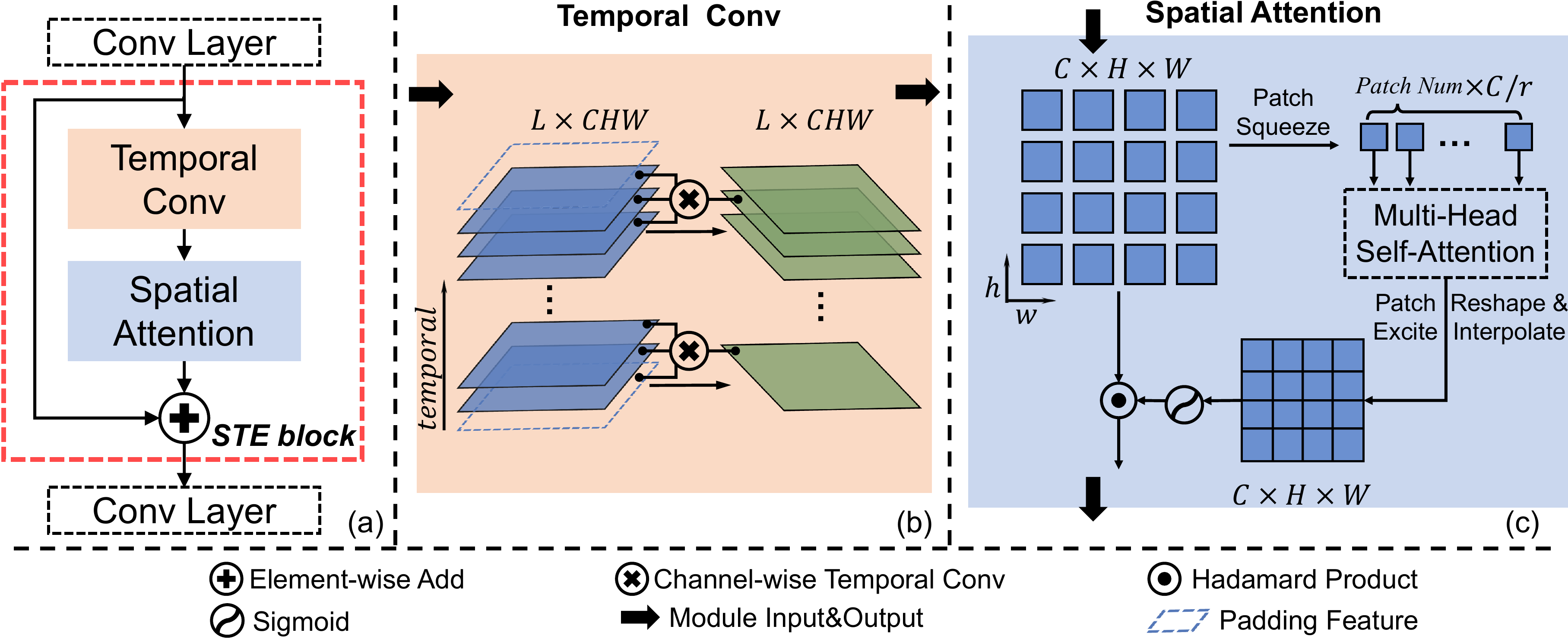}
\caption{Overview of the proposed STE block. \textbf{(a)}: The STE block is shown in the red dash box. \textbf{(b)(c)} depict the details of the two sub-blocks in STE.}
\label{fig:fig3}
\end{figure*}

Deepfake exhibits irregularities in both spatial and temporal space, so we specially incorporate the spatio-temporal designs in our model. Inspired by the motion modeling designs in \cite{jiang2019stm,li2020tea,liu2020teinet}, we propose the STE block to enhance the spatio-temporal learning of the CNN backbone.

The STE block is plugged between convolutional layers as shown in Fig.~\ref{fig:fig3}~(a). We feed the sequential spatial features into the Temporal Conv and Spatial Attention block to achieve spatio-temporal encoding. Given the features $\mathcal{F} \in \mathbb{R}^{C\times L\times H\times W}$, where $C$ is the channel dimension, $L$ is the temporal dimension, and $H\times W$ are the spatial dimensions, the Temporal Conv block imposes the channel-wise temporal convolution as:
\begin{equation}
\hat{\mathcal{F}}_{c,t,x,y} = \sum_{i\in[-1,0,1]} \mathrm{K}_{i}^{c}\cdot \mathcal{F}_{c,t+i,x,y},
\end{equation}
where $\mathrm{K}_{i}^{c}$ is the kernel of channel $c$, we implement this operation using 3D convolutional layer with the kernel size of $3\times 1\times 1$, i.~e., the features $\mathcal{F}_{t}$ at timestamp $t$ only interacts with its adjacent frames. 
In addition, the Spatial Attention block considers the spatial relations and imposes a patch-wise self-attention operation. Given $\hat{\mathcal{F}_t} \in \mathbb{R}^{C\times H\times W}$ at timestamp $t$, 
we first split the features at $H\times W$ dimensions into local patches using patch-convolutional operation, where the convolution kernel size and slide window stride are both set to $7\times 7$. Meanwhile, we squeeze the patch features at channel dimension with a reduction factor $r$ of 8. 
{\color{black}
The $\mathrm{Patch~Squeeze}$ (see Fig.~\ref{fig:fig3} (c)) operation can be denoted as:
\begin{equation}
\{\mathcal{S}_1,\mathcal{S}_2,...,\mathcal{S}_p\} = \mathrm{PatchSqueeze}(\hat{\mathcal{F}_t}),
\end{equation}
where $p=\lfloor H/7 \rfloor \times \lfloor W/7 \rfloor$ denotes the patch number, $\mathcal{S}_i \in \mathbb{R}^{C/r}$ is the output of the patch-convolution at a specific $7\times 7$ region.
}
Next, we feed the local patch features into a multi-head self-attention (``$\mathrm{SelfAtt}$") block \cite{vaswani2017attention})
{\color{black}
as:
\begin{equation}
\{\mathcal{S}'_1,\mathcal{S}'_2,...,\mathcal{S}'_p\}=\mathrm{SelfAtt}(\{\mathcal{S}_1,\mathcal{S}_2,...,\mathcal{S}_p\}).
\end{equation}
}
Thus, local patches can interact with the whole spatial space through this block. After that, the attention features are excited by a point-convolutional operation  (kernel size and slide window stride are set to $1\times 1$, ``$\mathrm{PointConv}$") to recover the original channel dimension 
{\color{black}
as:
\begin{equation}
\mathcal{S}''_i = \mathrm{PointConv}(\mathcal{S}'_i),~0< i\le p.
\end{equation}
}
Then we put the patch feature back to its original spatial location (see Fig.~\ref{fig:fig3} (c)) and interpolate the result to recover the original size. Finally, we have the Spatio-Temporal Enhanced feature as:
{\color{black}
\begin{equation}
\check{\mathcal{F}}_t = \hat{\mathcal{F}}_t \cdot \sigma(\operatorname{Interp-Reshape}(\{\mathcal{S}''_1,\mathcal{S}''_2,...,\mathcal{S}''_p\})),
\end{equation}
}
where $\sigma$ denotes sigmoid function and $\operatorname{Interp-Reshape}$ indicates interpolation after reshaping operation.

In this work, we plug the STE block before the first convolutional layer of every ResBlock in ResNet-34 \cite{he2016deep} as our model. 

\subsection{Detector Training and Discussion}
There are two basic training schemata depending on whether or not the \textit{fake} data is used. 1) Using only the \textit{real} data of existing deepfake dataset; 2) Both \textit{real} and \textit{fake} data are available. Regardless of the schema, the \textit{P-fake} Generator will create \textit{p-fake} on-the-fly to support the training. The label of our generated \textit{p-fake} is considered as the same as \textit{fake} data. As a binary classification problem, our model is trained with 0/1 label (0 for \textit{real}, 1 for \textit{fake} and \textit{p-fake}) supervision using binary cross-entropy loss:
\begin{equation}
\mathcal{L}(\mathbb{W}) = -y_{gt}{\rm log}\left (y_{pred}   \right )-(1-y_{gt}){\rm log}\left (1-y_{pred}\right ), \label{bce_loss}\\
\end{equation}
where $\mathbb{W}$ is the trainable parameters in our network, $y_{gt}$ denotes the ground truth label, and $y_{pred}$ is the predicted confidence score.

With the \textit{p-fake} support in training, our method shows several fabulous properties for deepfake detection:
{\color{black}
\textbf{1)} Deep Neural Networks can be trained without requiring labor-intensive efforts. 
The training of Deepfake detection mostly relies on publicly available deepfake methods which are limited in specific domains.
Additionally, annotating in-the-wild fake videos accurately is challenging given the increasing fidelity of deepfake generation techniques. For example, we can hardly recognize fake video footage\footnote{https://www.youtube.com/watch?v=iyiOVUbsPcM} without title information. 
}
\textbf{2)} The \textit{real} data is almost inexhaustible. Thus, with our proposed \textit{P-fake} Generator, in theory, we have an infinite variety of data for training.
\textbf{3)} Alternatively, combined with existing deepfake datasets, our proposed \textit{P-fake} Generator can work as a plug-and-play module to enhance these datasets and be applied to all the existing learning-based methods.
\section{Experiments}
In this section, we first briefly introduce the experiment setups, then present abundant experimental results to demonstrate the SOTA performance of our method, and finally validate the effectiveness of our designs through ablations.

\noindent\textbf{Datasets.}
We conduct our experiments on several recently published datasets.
1) FaceForensics++ (FF++) \cite{rossler2019faceforensics++} is the most popular dataset used to evaluate deepfake detection performance. It contains 1000 real videos collected from the Internet, and 4000 fake videos generated by four kinds of deepfake techniques including Deepfakes (DF) \cite{deepfakes2022ff++}, FaceSwap (FS) \cite{faceswap2022ff++}, Face2Face (F2F) \cite{thies2016face2face}, and NeuralTextures (NT) \cite{thies2019deferred}. In addition, FF++ provide three video qualities (raw, c23 for visually nearly lossless compression, and c40 for heavy compression) to evaluate detection methods against real-world streaming media compression. 
{\color{black}
All real and fake videos are split into train/val/test sets according to a ratio of 72/14/14. Our models are trained solely on this dataset for the following experiments.
}
2) WildDeepfake (Deepwild) \cite{zi2020wilddeepfake} contains 3805 real face sequences and 3509 fake face sequences. Especially, all these sequences are manually collected from the Internet, which better supports deepfake detection against real-world scenarios. 
{\color{black}
We use the test set of this dataset for cross-dataset evaluation, which contains 396 real sequences and 410 fake sequences.
}
3) DeepFake Detection Challenge preview (DFDCP) \cite{dolhansky2019deepfake} is one of the most challenging datasets with deepfakes generated from the original videos filmed in a variety of environments. 
{\color{black}
We use the test set containing 276 real videos and 501 fake videos.
}
4) Celeb-DF (CDF) \cite{li2020celeb} contains high-quality face-swapped videos, especially, the authors use a Kalman smoothing algorithm to reduce imprecise variations of facial landmarks in each frame, which greatly relieves the temporal incoherence and makes this dataset more challenging. 
{\color{black}
Test set of CDF containing 178 real videos and 340 fake videos is used for cross-dataset evaluation.
}
5) DeepfakeDetection (DFD) \cite{dfd_dataset} is a dataset contains more than 3000 videos released by Google to support deepfake detection research. 
{\color{black}
All of the videos are used for testing in our experiments.
}
6) {Face Shifter} (FSh) provides high-quality face-swapped videos based on the 1000 real videos in FF++ using the SOTA forgery method \cite{li2020advancing}. 
{\color{black}
The test set, containing 140 real videos and 140 fake videos, is used for evaluation.
}

\noindent\textbf{Evaluation Metrics.}
Following the previous works \cite{li2020face,sun2021dual,zhao2021learning}, we use 
Accuracy score (ACC), Area Under the Receiver Operating Characteristic Curve (AUC), and Equal Error Rate (EER) as our evaluation metrics. 

\noindent\textbf{Data Preprocessing.}
Except for Deepwild (which provides face crops rather than videos), we first extract frames from videos and crop the face regions using MTCNN \cite{zhang2016joint}. After that, face crops are resized to $299\times 299$. For the \textit{real} data used to generate \textit{p-fake}, we extract the 68-points facial landmark using Dlib \cite{king2009dlib}.

\noindent\textbf{Training Strategy.}
We train our model using Adam optimizer with an initial learning rate $2\times 10^{-4}$ and weight decay $10^{-7}$. If the validation loss does not decline for 20 epochs, we decay the learning rate by the factor of 0.3, so that our model can converge after several decays. 

\noindent\textbf{Implementation Details.}
We construct our model based on the CNN backbone, ResNet-34 \cite{he2016deep} and we plug the proposed STE block before the first convolutional layer of each ResBlock. For the input data, our model takes sequences with 32 successive frames. Common data augmentations including horizontal flip, random rotate, and image compression are applied. All the experiments are conducted on the NVIDIA 3090 GPU.

\begin{table}[!b]
\begin{center}
\caption{Performance evaluated on FF++ (raw). $\dagger$:~re-implemented using the official code by ourself.}
\begin{tabular}{|l|c|ccccc|}
\hline
\multirow{2}{*}{Method} & \multirow{2}{*}{\begin{tabular}[c]{@{}c@{}}Train\\ Set\end{tabular}} & \multicolumn{5}{c|}{Test Set (AUC\%$\uparrow$)}                 \\
\cline{3-7}
  & & DF & F2F & FS & NT & \textbf{\textit{Avg}} \\
\hline
Xception \cite{chollet2017xception} & \multirow{1}{*}{FF++}  & 99.38 & 99.53 & 99.36 & 97.29 & 98.89 \\
\hline
Face X-ray \cite{li2020face} & \multirow{4}{*}{\begin{tabular}[c]{@{}c@{}}FF++\\ only \\ \textit{real}\end{tabular}} & 99.17     & 98.57     & 98.21    & {98.13}          & {98.52}   \\
PCL+I2G \cite{zhao2021learning} & & {100.00}    & {98.97}     & {99.86}    & 97.63          & {99.11}   \\
EB4+SBI \cite{shiohara2022detecting} $\dagger$ & & {99.99}    & {99.79}     & {99.58}    & 97.81          & {99.29} \\
\textbf{Ours} & & {99.76}     & {99.68}     & {99.65}   & {99.40}          & \textbf{99.62}   \\
\hline
\end{tabular}
\label{tab:FF self-supervised}
\end{center}
\end{table}

\begin{table}[!b]
\begin{center}
\caption{Performance evaluated on FF++ (c23, compressed). $\dagger$:~re-implemented using the official code by ourself.}
\begin{tabular}{|l|c|ccccc|}
\hline
\multirow{2}{*}{Method} & \multirow{2}{*}{\begin{tabular}[c]{@{}c@{}}Train\\ Set\end{tabular}} & \multicolumn{5}{c|}{Test Set (AUC\%$\uparrow$)}                 \\
\cline{3-7}
  & & DF & F2F & FS & NT & \textbf{\textit{Avg}} \\
\hline
Face X-ray \cite{li2020face} & \multirow{3}{*}{\begin{tabular}[c]{@{}c@{}}FF++\\ only\\ \textit{real}\end{tabular}} & -     & -    & -    & -          & {87.35}   \\
EB4+SBI \cite{shiohara2022detecting} $\dagger$ & & {97.53}    & {88.99}     & {96.42}    &  82.80          & {91.44} \\
\textbf{Ours} & & {98.88}     & {98.13}     & {94.13}   & {88.38}          & \textbf{94.88}   \\
\hline
\end{tabular}
\label{tab:FF c23 self-supervised}
\end{center}
\end{table}

\setlength{\tabcolsep}{8pt}
\begin{table*}[b]
\begin{center}
\caption{Cross-dataset evaluation. We train our model using only \textit{real} videos in FF++ and tested on DFD, DFDCP, Deepwild and CDF, respectively. We report the performance of AUC\%$\uparrow$ and EER\%$\downarrow$, with the best in bold and the second best underlined. $\dagger$:~re-implemented using the official code by ourself.}
\begin{tabular}{|l|c|cccccccc|}
\hline
\multirow{2}{*}{Method} & \multirow{2}{*}{\begin{tabular}[c]{@{}c@{}}Train\\ Set\end{tabular}} & \multicolumn{2}{c}{DFD} & \multicolumn{2}{c}{DFDCP} & \multicolumn{2}{c}{Deepwild} & \multicolumn{2}{c|}{CDF} \\ \cline{3-10} 
             &  & AUC\%   & EER\%   & AUC\%   & EER\%   & AUC\%   & EER\%   & AUC\%   & EER\%   \\ \hline
F3-Net \cite{qian2020thinking}  &  \multirow{9}{*}{\begin{tabular}[c]{@{}c@{}}FF++\end{tabular}}     & 86.10 & 26.17 & 72.88 & 33.38 & 67.71 & 40.17 & 71.21 & 34.03 \\
MAT \cite{zhao2021multi}    &        & 87.58 & 21.73 & 67.34 & 38.31 & 70.15 & 36.53 & 76.65 & 32.83 \\
GFF \cite{luo2021generalizing}  &         & 85.51 & 25.64 & 71.58 & 34.77 & 66.51 & 41.52 & 75.31 & 32.48 \\
LTW \cite{sun2021domain}   &  & 88.56 & 20.57 & 74.58 & 33.81 & 67.12 & 39.22 & 77.14 & 29.34 \\
{\color{black} LipForensics} \cite{haliassos2021lips} $\dagger$ &  & 75.27 & 34.16 & 67.17 & 39.18 & 66.14 & 37.80 & 72.49 & 34.09 \\
FTCN-TT \cite{zheng2021exploring}  &     & - & - & 74.00 & - & - & -& {86.90} & - \\
Local-relation \cite{chen2021local} & & 89.24 & 20.32 & 76.53 & 32.41 & 68.76 & 37.50 & 78.26 & 29.67 \\
DCL \cite{sun2021dual}  &         & 91.66 & 16.63 & {76.71} & {31.97} & {71.14} & {36.17} & 82.30 & {26.53} \\
\hline
Face X-ray \cite{li2020face}&  \multirow{5}{*}{\begin{tabular}[c]{@{}c@{}}FF++\\ only\\ \textit{real}\end{tabular}}   & {93.47} & {12.72} & 71.15  & 32.62  & -  & - & 74.76 & 31.16  \\ 
PCL+I2G \cite{zhao2021learning}& & \textbf{99.07} & - & 74.37  & -  & -  & - & \underline{90.03} & -  \\ 
EB4+SBI \cite{shiohara2022detecting} $\dagger$ & & 97.70 & \underline{7.16} & \textbf{85.12}  & \textbf{23.19}  & \underline{71.18}  & \underline{34.15} & {89.86} & \underline{18.82}  \\ 
R34+SBI \cite{shiohara2022detecting} $\dagger$ & & 93.55 & 13.10 & 81.24 & 26.09  & 58.03  & 44.39 & 83.79 & 23.82  \\ 
\textbf{Ours}   &       & \underline{98.25}   & \textbf{6.30}  & \underline{85.01} & \underline{23.74}  &   \textbf{81.85}   &  \textbf{26.59}     & \textbf{90.17}  &  \textbf{17.51}    \\ \hline
\end{tabular}
\label{tab:cross-datasets}
\end{center}
\end{table*}

\begin{table*}[h]
\begin{center}
\caption{Performance gain of \textit{p-fake} evaluated on FF++ (c23), FSh, DFDCP and CDF. We report the performance of AUC\%$\uparrow$ and EER\%$\downarrow$.}
\begin{tabular}{|l|cccccccc|}
\hline
  \multirow{2}{*}{Train Data} &
  \multicolumn{2}{c}{FF++} &
  \multicolumn{2}{c}{FSh} &
  \multicolumn{2}{c}{DFDCP} &
  \multicolumn{2}{c|}{CDF} \\ \cline{2-9} 
                  & AUC\% & EER\% & AUC\% & EER\% & AUC\% & EER\% & AUC\% & EER\% \\
\hline
\textit{real}+\textit{fake}   & 99.12   & 2.86     & 86.36  & 20.71 & 73.20  & 32.37       & 77.80   & 30.59      \\
\textit{real}+\textit{p-fake} & 94.88   & 13.21  & 99.15  & 3.57   & \textbf{85.01}    & \textbf{23.74}  & \textbf{90.17}  & \textbf{17.51} \\
\textit{real}+\textit{fake}+\textit{p-fake}    &     \textbf{99.33}  & \textbf{1.79}   & \textbf{99.61} & \textbf{2.86} & 83.51 & 24.51 & 85.68 & 21.47    \\
\hline
\end{tabular}
\label{tab:pfake_ablation}
\end{center}
\end{table*}

\begin{table}[h]
\caption{In-dataset evaluation on FF++ with different compression qualities. Our method shows better performance (ACC\%$\uparrow$ and AUC\%$\uparrow$) on lower-quality videos, where the best is shown in bold text and the second best is underlined.}
\begin{center}
\begin{tabular}{|l|cccc|}
\hline
\multirow{2}{*}{{Method}} & \multicolumn{2}{c}{c23} & \multicolumn{2}{c|}{c40} \\ \cline{2-5} 
 & ACC\%   & AUC\%   & ACC\%   & AUC\%   \\ 
\hline
Add-Net \cite{zi2020wilddeepfake}              & 96.78 & 97.74 & 87.50 & 91.01 \\
Two-Branch \cite{masi2020two}          & 96.43 & 98.70 & 86.34 & 86.59 \\
MAT \cite{zhao2021multi}     & \underline{97.60} & 99.29 & 88.69 & 90.40 \\
FDFL \cite{li2021frequency}              & 96.69 & \underline{99.30} & \underline{89.00} & \underline{92.40} \\
\textbf{Ours}  & \textbf{98.29} & \textbf{99.33} & \textbf{90.47}  & \textbf{93.71} \\
\hline
\end{tabular}
\end{center}
\label{tab:ff_indataset}
\end{table}

\setlength{\tabcolsep}{6.5pt}
\begin{table}[]
\caption{Performance evaluated on FSh and CDF. Two numbers in each cell denote test results of model trained w/o and w/ \textit{p-fake}.}
\begin{center}
\begin{tabular}{|ll|ccc|}
\hline
\multicolumn{2}{|l|}{Model~$\rightarrow$} & Xception \cite{chollet2017xception} & Swin \cite{liu2021swin}  & TSM-R34 \cite{lin2019tsm} \\ \hline
\multicolumn{1}{|l|}{\multirow{2}{*}{AUC\%$\uparrow$}} & FSh & 78.60/95.84 & 84.21/92.90 &  81.73/96.68  \\
\cline{2-5}
\multicolumn{1}{|l|}{}                     & CDF  & 74.90/82.50 & 81.79/84.75     &  75.42/82.38  \\ \hline
\end{tabular}
\end{center}
\label{tab:ablation_pfake+backbones}
\end{table}

\setlength{\tabcolsep}{6pt}
\begin{table*}[!h]
\begin{center}
\caption{Performance of AUC\%$\uparrow$ on four subsets of FF++ (raw) with different editing methods, different masks, and different blending methods in the \textit{P-fake} Generator. ``only \textit{face}" indicates that we use only the whole face mask. ``only \textit{alpha}" denotes that we use only alpha blending in the Blender.}
\begin{tabular}{|ccc|c|c|cccc|}
\hline
\multicolumn{3}{|c|}{Editing Method} &
  \multirow{2}{*}{\begin{tabular}[c]{@{}c@{}}Random\\ Mask\end{tabular}} &
  \multirow{2}{*}{\begin{tabular}[c]{@{}c@{}}Random\\ Blending\end{tabular}} &
  \multicolumn{4}{c|}{AUC\%} \\ \cline{1-3} \cline{6-9} 
Photometric & Geometric & Frequency &  &  & DF & F2F & FS & NT \\
\hline
\checkmark            &           & & \checkmark & \checkmark & 98.19  & 96.88  & 94.06 & 95.58   \\
            &\checkmark           & & \checkmark & \checkmark & 98.48 & 95.53  & 93.37 & 93.42\\
            &           &\checkmark & \checkmark & \checkmark & 96.96   & {91.21}   &  78.06 & 94.22    \\
\hline
\checkmark  &\checkmark &\checkmark & only \textit{face} & \checkmark  & 99.73    & 97.63    &  80.66  & 97.76  \\
\checkmark  &\checkmark &\checkmark & \checkmark & only \textit{alpha}  & 99.25   & 98.92    & 96.98    & 98.08   \\
\hline
\checkmark  &\checkmark &\checkmark & \checkmark & \checkmark  &  99.76  & 99.68 & 99.65  & 99.40  \\
\hline
\end{tabular}
\label{tab:editing_ablation}
\end{center}
\end{table*}

\begin{table}[!h]
\centering
\caption{Ablation on the STE block. We train the models on FF++ (c23) using \textit{real}, \textit{p-fake} data and report the AUC\%$\uparrow$ performance tested on DFDCP, CDF and FSh. Without STE, our model remains only spatial modeling ability.} 
\begin{tabular}{|l|ccc|}
\hline
Model        & DFDCP & CDF & FSh \\
\hline
TSM-R34 \cite{lin2019tsm}  & 76.68  & 82.38 & 96.68 \\
SlowFast-R50 \cite{feichtenhofer2019slowfast}  & 71.34 & 78.73 & 95.99   \\
Ours w/o STE           & 76.04  & 81.89    & 94.11    \\
STE w/o TempConv     & 77.97  & 85.02  &   95.94  \\
STE w/o SpatiAtt     & 79.56  & 84.56  &   97.59 \\
\textbf{Ours}         & 85.01    & 90.17    & 99.15   \\
\hline
\end{tabular}
\label{tab:STE ablation} 
\end{table}

\begin{table}[]
\caption{Performance evaluated on FSh and CDF. Models are trained on FF++ (c23) without aids of \textit{p-fake}.}
\begin{center}
\begin{tabular}{|ll|ccc|}
\hline
\multicolumn{2}{|l|}{Model~$\rightarrow$}                       & R34 \cite{he2016deep} & TSM-R34 \cite{lin2019tsm} & \textbf{Ours} (R34+STE) \\ \hline
\multicolumn{1}{|l|}{\multirow{2}{*}{AUC\%$\uparrow$}} & FSh & 72.33 & 81.73 & 86.36        \\
\cline{2-5}
\multicolumn{1}{|l|}{}                     & CDF  & 73.13  & 75.42 &  77.80       \\ \hline
\end{tabular}
\end{center}
\label{tab:ablation_STE_no_pfake}
\end{table}

\subsection{Comparing with Previous Methods}
As the goal we mentioned previously, to achieve deepfake detection, we need only the \textit{real} data, and \textit{P-fake} Generator will provide the \textit{p-fake} as the negative sample. 

\noindent\textbf{In-Dataset Performance on FF++.} 
We first conduct the experiment on FF++ to demonstrate the effectiveness of our main idea. 
{\color{black}
Using only the 720 \textit{real} videos from the train set, we trained our model on FF++ and evaluated it on four subsets.
}
The results are shown in the Table~\ref{tab:FF self-supervised}.
\textcolor{black}{
Xception \cite{chollet2017xception} represents the commonly used baseline in the field of deepfake detection, which predicts the fake probability by averaging the outcomes of individual frames. When training and testing are performed on data with the same distribution of forgery cues, good performance is naturally achieved by learning-based methods, e.g., Xception. Note only this baseline is trained using both \textit{real} and \textit{fake} videos of the train set.
However, other methods in Table~\ref{tab:FF self-supervised}, use only a subset of the train set (only the \textit{real} part) and do not rely on fitting forgery features of existing deepfakes. The better performance of ours demonstrates the effectiveness of our approach.
In our later experiments, we find that incorporating \textit{fake} data from the train set can further improve the performance under this in-dataset evaluation.
}
Nonetheless, we notice the performances evaluated on the raw version of FF++ turn to saturation. We thereupon further compare with recent arts on the compressed c32 version in Table~\ref{tab:FF c23 self-supervised}. When the video is compressed, specific forgery traces are less significant in the spatial dimension, thus Face X-ray \cite{li2020face} and SBI \cite{shiohara2022detecting} both demonstrate sub-optimal results. While our method consider the overall spatio-temporal regularity, thus maintaining better performance even the forgery traces at spatial dimension are less significant.
Performances within FF++ validate the feasibility of our main idea, while generalizability remains a major problem of existing deepfake detection methods, as the performance is not guaranteed when testing on deepfakes generated by unseen techniques. Consider further the reality of the arms race between detectors and forgers, generalizability is an important criterion to measure the effectiveness of a detection method in the real world.

\noindent\textbf{Cross-Dataset Performance.}
Considering the more challenging cross-dataset setting, we further evaluate our method on DFD, DFDCP, Deepwild, and CDF. Following the setup in the SOTA method \cite{sun2021dual}, we train our model on the c23 version of FF++ and test on other four datasets 
{\color{black}
(using the official split of the test sets as we described in the beginning of this section).
}
{\color{black}
Note that our method, Face X-ray \cite{li2020face}, PCL+I2G \cite{zhao2021learning}, and SBI \cite{shiohara2022detecting} use only the 720 \textit{real} data of FF++ in training. While other learning-based methods are trained using 720 \textit{real} and 2880 \textit{fake} videos.
}
The results are shown in Table~\ref{tab:cross-datasets}, where our method presents superior performance on all four datasets.
As only the face-blending artifacts are considered in Face X-ray, it shows relatively inferior performance. Comparing with PCL+I2G, which models the face-swapping inconsistency in spatial dimension to identify deepfakes, our method outperform it on DFDCP with a clear margin. Moreover, SBI is a concurrent work that proposes a similar blending-after-editing pipeline as our \textit{p-fake} Generator. It achieves similar performance with a stronger backbone (EfficientNet-B4 \cite{tan2019efficientnet}), while using the same backbone, ResNet-34 (R34) \cite{he2016deep}, our model shows better performance attributing to the fundamental spatio-temporal regularity disruption spotting.
On the most challenging Deepwild, our method surpasses the SOTA method by about 10 percentage points in terms of AUC\%.
We think this is due to the large diversity of deepfakes in Deepwild, which makes other methods fail to generalize well from seen deepfakes.

\subsection{Ablation Study}
\label{sec:ablation}
\noindent\textbf{Performance Gain of \textit{p-fake}.}
To clearly demonstrate the effect of the proposed \textit{p-fake}, we train our model using different data combinations on FF++. Both the in-dataset and cross-dataset results are shown in Table~\ref{tab:pfake_ablation}. Although better in-dataset performance is reported using ``\textit{real}+\textit{fake}", model trained using ``\textit{real}+\textit{p-fake}" shows significantly better performance on unseen datasets. As the AUC\% is improved from 86.36\% to 99.15\%, from 77.80\% to 90.17\% on FSh and CDF, respectively.
When combining the three kinds of data together, the performance on FF++ is further improved, we leave more in-dataset results of ``\textit{real}+\textit{fake}+\textit{p-fake}" combination in the next paragraph.
{\color{black}
Another observation is that adding fake data from FF++ to the training set resulted in a significant drop in performance on DFDCP and CDF, while the performance on FF++ and FSh datasets increased. This suggests that including more fake data with different forgery patterns in training (i.e., \textit{fake} of FF++ train set) may not necessarily benefit discovering of out-of-domain forgeries (i.e., test set of DFDCP and CDF). 
}

\begin{figure*}[b]
\centering
\includegraphics[width=\linewidth]{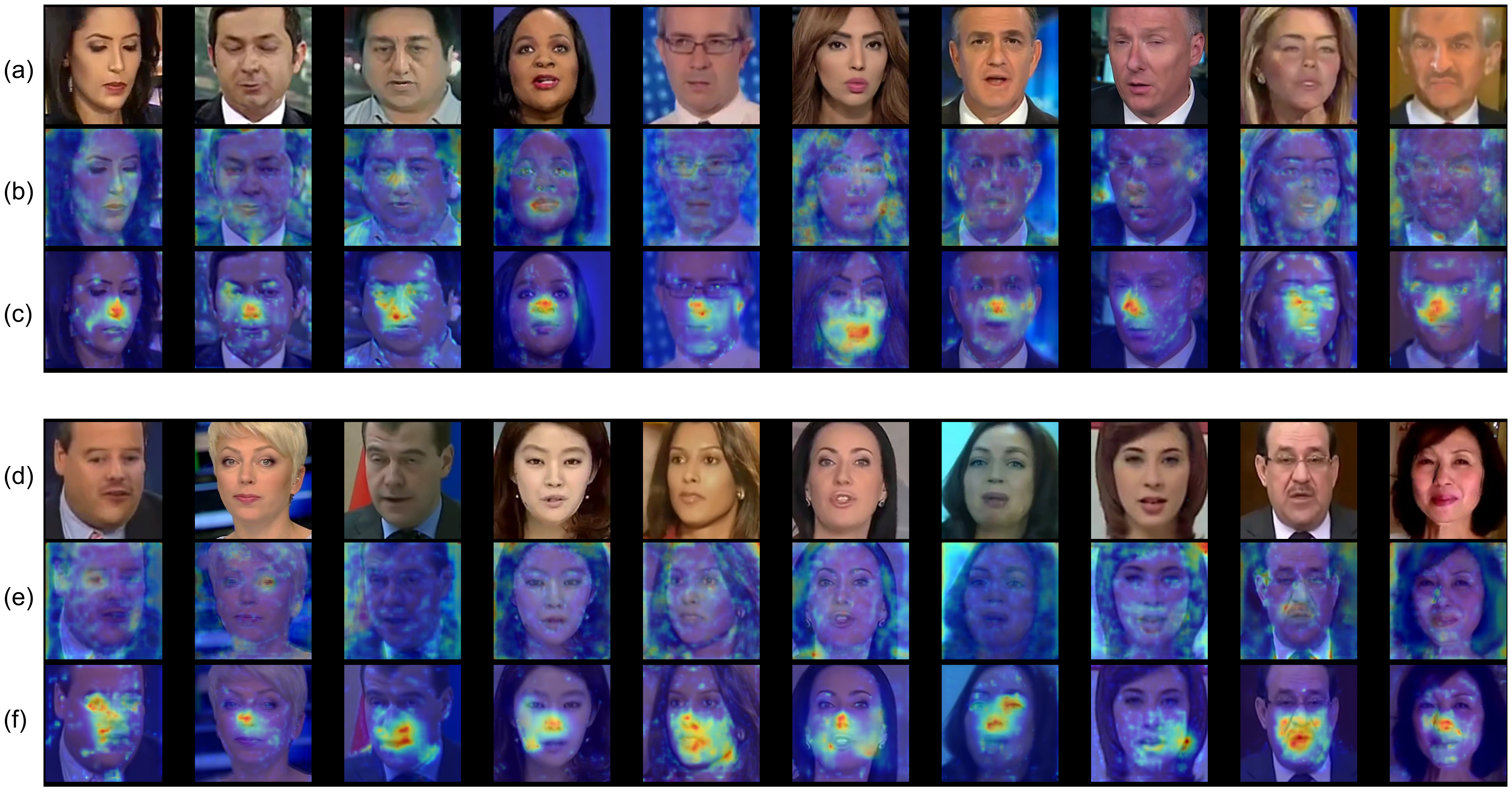}
\caption{{(a),(d)}: The original frames. {(b),(e)}: CAMs of our model trained using ``\textit{real}+\textit{p-fake}" data. {(c),(f)}: CAMs of the model trained without \textit{p-fake}}
\label{fig:fig5}
\end{figure*}

\noindent\textbf{``\textit{real}+\textit{fake}+\textit{p-fake}" Combination Evaluation.}
In-dataset evaluation on FF++ is widely adopted by many previous methods. 
Here we combine the self-created \textit{p-fake} with \textit{real} and \textit{fake} data of FF++ (``\textit{real}+\textit{fake}+\textit{p-fake}") in training, and test our model on different compressed versions of FF++, respectively. 
The results are shown in Table~\ref{tab:ff_indataset}.
{\color{black}
Note we exclude the raw version here since even a simple baseline such as Xception can achieve saturated performance on the raw data (see Table \ref{tab:FF self-supervised}), which hardly reveals differences among methods.
}
Our method achieves the best performance on both the compressed versions. Similar results are reported by F3-Net, while they explore the subtle frequency forgery clues with more complex model ($\sim$41.6M parameters). In contrast, we achieve the SOTA performance with about half the parameters ($\sim$23.5M), benefiting from the discriminatory irregularity learning aided by the \textit{P-fake} Generator and spatio-temporal encoding of the STE block. 

{\color{black}
\noindent\textbf{Does \textit{p-fake} benefit different models?}
To examine whether the performance improvement achieved with the use of \textit{p-fake} is transferable to existing methods, we conduct additional experiments on three types of learning-based models: the commonly used CNN backbone Xception \cite{chollet2017xception}, the popular Transformer-based \cite{dosovitskiy2020image} model Swin \cite{liu2021swin}, and a temporal-aware model TSM-R34 \cite{lin2019tsm}.
Three models are train on FF++ using data combinations of ``\textit{real}+\textit{fake}'' (denoted w/o \textit{p-fake} and ``\textit{real}+\textit{p-fake}'' (denoted w/ \textit{p-fake}).
As the results shown in Table \ref{tab:ablation_pfake+backbones}, we see a clear promotion is obtained for all the models when \textit{p-fake} is included. 
We also observed that the Transformer-based \cite{dosovitskiy2020image} model achieves the best performance without aids of \textit{p-fake}. 
This may suggest that the patch-based learning could be more effective for the task of deepfake detection, as most of the menacing deepfakes involve partial modifications such as face swaps and lip syncing. 
This finding is also supported by our later ablation of the patch-level-designed STE.
}

\noindent\textbf{Diversity of \textit{p-fake}.}
In the \textit{P-fake} Generator, we consider editing methods from different aspects, different disruption regions, and different blending methods to cover larger irregularity distribution. Here we validate these designs in FF++ and show the results in Table~\ref{tab:editing_ablation}. It shows that diverse heterogeneous features introduced by editing methods at different perspectives result in better performance. In combination with the random mask and blending, our designs consistently improve the final performance.

\noindent\textbf{Benefits of STE.}
In this part, we conduct ablation study on the effect of our proposed STE block. As we insert the STE block before the ResBlock in ResNet-34 to enhance the spatio-temporal modeling, by removing SET blocks, the model presents only spatial encoding ability. We illustrate the empirical results in Table~\ref{tab:STE ablation}. With the unified spatio-temporal modeling, our model shows non-trivial improvements, especially on FSh and CDF, the average AUC\% is improved from 88.00\% to 94.66\%. 
We also compare with two spatio-temporal baselines, TSM-R34 \cite{lin2019tsm} and SlowFast-R50 \cite{feichtenhofer2019slowfast}, which are firstly proposed for action recognition. The results indicate that our elaborately devised STE block is more suitable for the task of deepfake detection as more fine-grained patch-level modeling is considered. 

{\color{black}
\noindent\textbf{Does STE contribute w/o \textit{p-fake}?}
We further validate the effectiveness of STE by conducting comparison without aids of \textit{p-fake}. We train three models on training set of FF++ and test on FSh and CDF. The results are tabulated in Table \ref{tab:ablation_STE_no_pfake}. Both TSM-R34 \cite{lin2019tsm} and our model are built based on ResNet (R34) \cite{he2016deep}, but the proposed STE considers more fine-grained patch-level spatio-temporal modeling shows more powerful forgery spotting ability. Comparing with R34, the performance is improved from 72.73\% to 82.08\% averaged from FSh and CDF.
}

\subsection{Activation Visualization}
To intuitively demonstrate different patterns learned with \textit{p-fake} support, we compare the CAM~\cite{selvaraju2017grad} visualizations between models trained with and without \textit{p-fake} data. As shown in Fig.~\ref{fig:fig5}, our model focuses on the overall regularity under the supervision of \textit{p-fake}. While the model trained without \textit{p-fake} only pays attention to the most semantically suspicious facial features.
{
\color{black}
Based on the results of our quantitative experiments, we believe that paying global attention to both the background and facial parts can improve the generalization ability of deepfake detectors.
}

\section{Conclusions}
Jointly considering spatio-temporal properties of real videos, we introduce a plug-and-play module, \textit{P-fake} Generator, to create \textit{p-fake} videos as negative samples for training.
Such practice brings a broad range of possible regularity disruptions and greatly improve the generalizability of the detector.
Based on our \textit{P-fake} Generator, we further present a specially devised STE block to better capture the spatio-temporal irregularity patterns.
Extensive experimental results demonstrate the superior performance of our method. 
We hope that the idea of regularity disruption spotting can be incrementally studied to better deal with the real-world demand of deepfake detection.

\bibliographystyle{IEEEtran}
\bibliography{IEEEabrv,egbib}

\end{document}